\documentclass[letterpaper]{article} % DO NOT CHANGE THIS
\usepackage{aaai2026}  % DO NOT CHANGE THIS
\usepackage{times}  % DO NOT CHANGE THIS
\usepackage{helvet}  % DO NOT CHANGE THIS
\usepackage{courier}  % DO NOT CHANGE THIS
\usepackage[hyphens]{url}  % DO NOT CHANGE THIS
\usepackage{graphicx} % DO NOT CHANGE THIS
\usepackage{subfigure}
\usepackage{xcolor}
\usepackage{amssymb}
\usepackage{float}
\usepackage{amsmath}
\usepackage{natbib}  % DO NOT CHANGE THIS AND DO NOT ADD ANY OPTIONS TO IT
\usepackage{caption} % DO NOT CHANGE THIS AND DO NOT ADD ANY OPTIONS TO IT
\usepackage{algorithm}
\usepackage{algorithmic}

\usepackage{newfloat}
\usepackage{listings}
\DeclareCaptionStyle{ruled}{labelfont=normalfont,labelsep=colon,strut=off} % DO NOT CHANGE THIS
\floatstyle{ruled}
\newfloat{listing}{tb}{lst}{}
\floatname{listing}{Listing}
\title{OptiCorNet: Optimizing Sequence-Based Context Correlation for Visual Place Recognition}
\author{
    Zhenyu Li\textsuperscript{\rm 1, *}, Tianyi Shang\textsuperscript{\rm 2}, Pengjie Xu\textsuperscript{\rm 3}, Ruirui Zhang\textsuperscript{\rm 1}, Fanchen Kong\textsuperscript{\rm 1}\\
        }
\affiliations{
    \textsuperscript{\rm 1}Qilu University of Technology (Shandong Academy of Sciences)\\
    \textsuperscript{\rm 2}Fuzhou University\\
    \textsuperscript{\rm 3}Shanghai Jiao Tong University\\
}

\usepackage{bibentry}
\begin{document}
\nocopyright
\maketitle

\begin{abstract}
Visual Place Recognition (VPR) in dynamic and perceptually aliased environments remains a fundamental challenge for long-term localization. Existing deep learning-based solutions predominantly focus on single-frame embeddings, neglecting the temporal coherence present in image sequences. This paper presents OptiCorNet, a novel sequence modeling framework that unifies spatial feature extraction and temporal differencing into a differentiable, end-to-end trainable module. Central to our approach is a lightweight 1D convolutional encoder combined with a learnable differential temporal operator, termed Differentiable Sequence Delta (DSD), which jointly captures short-term spatial context and long-range temporal transitions. The DSD module models directional differences across sequences via a fixed-weight differencing kernel, followed by an LSTM-based refinement and optional residual projection, yielding compact, discriminative descriptors robust to viewpoint and appearance shifts. To further enhance inter-class separability, we incorporate a quadruplet loss that optimizes both positive alignment and multi-negative divergence within each batch. Unlike prior VPR methods that treat temporal aggregation as post-processing, OptiCorNet learns sequence-level embeddings directly, enabling more effective end-to-end place recognition. Comprehensive evaluations on multiple public benchmarks demonstrate that our approach outperforms state-of-the-art baselines under challenging seasonal and viewpoint variations. The code will be publicly released.
\end{abstract}
% Uncomment the following to link to your code, datasets, an extended version or similar.
% You must keep this block between (not within) the abstract and the main body of the paper.
% \begin{links}
 %    \link{Code}{https://github.com/CV4RA/OptiCorNet}
 %    \link{Datasets}{https://aaai.org/example/datasets}
 %    \link{Extended version}{https://aaai.org/example/extended-version}
 %\end{links}
\section{Introduction}
\indent Visual Place Recognition (VPR) is fundamental to visual navigation, mapping, and localization, allowing a robot or agent to identify previously visited locations based solely on visual observations \cite{li2025place}. It is a crucial element of long-term autonomous systems, particularly in the face of changing environmental conditions, such as variations in illumination, weather, structural dynamics, and viewpoint transformations \cite{li2025multi}. Despite substantial advancements in deep learning-based visual recognition, achieving robust and scalable place recognition continues to be a significant and challenging problem.\\
\indent Early VPR systems relied on hand-crafted features such as SIFT, SURF, and GIST. While these methods are computationally efficient, they are vulnerable to perceptual aliasing and environmental dynamics \cite{lowry2015visual}. The advent of deep learning, particularly through Convolutional Neural Networks (CNNs), has led to a paradigm shift in the extraction of visual descriptors. Techniques such as NetVLAD \cite{arandjelovic2016netvlad}, Patch-Netvlad \cite{hausler2021patch}, and DELG \cite{cao2020unifying} leverage CNN backbones to extract both global and local features, which are subsequently aggregated into robust image-level descriptors. These approaches demonstrate strong invariance to moderate changes in appearance and viewpoint. \\
\begin{figure}
    \centering
    \includegraphics[width=0.85\linewidth]{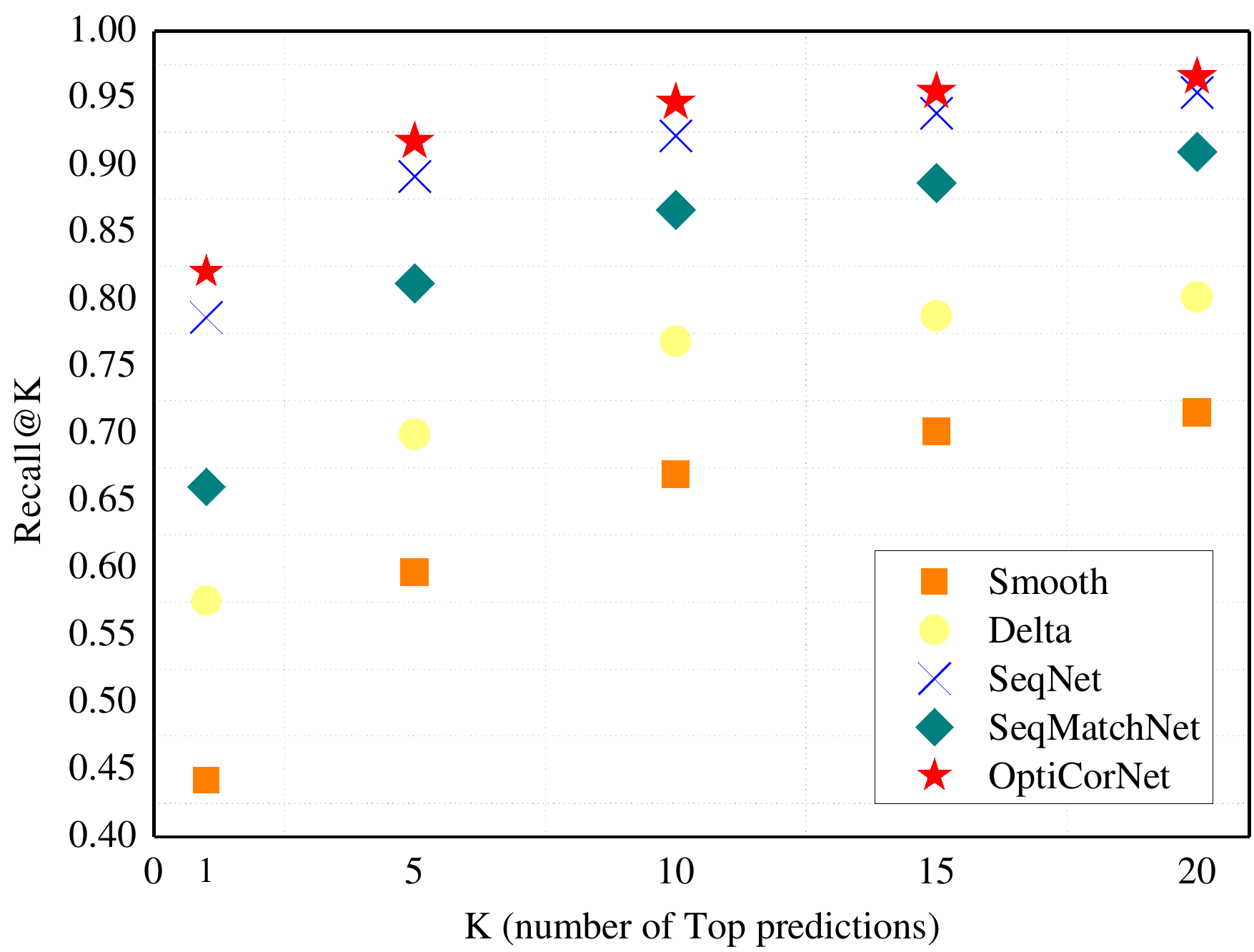}
    \caption{Performance comparison between the proposed OptiCorNet and existing sequence-based methods on Nordland datasets.}
    \label{fig1}
\end{figure}
\indent Recent efforts have investigated the potential of Transformer-based architectures for VPR, capitalizing on their global self-attention mechanisms and ability to model long-range dependencies. Vision Transformers (ViTs), such as DINOv2 \cite{oquab2023dinov2}, have shown impressive generalization capabilities in both classification and retrieval tasks by attending to spatially distributed visual tokens. When adapted for place recognition, these models can encode richer semantic context compared to conventional CNNs.\\
\indent However, the majority of deep VPR models treat each input frame as an isolated observation, extracting embeddings from individual images without considering the sequential nature of visual navigation \cite{li2024feature}, \cite{li2024toward}. This per-frame modeling approach results in suboptimal performance in scenarios characterized by significant visual ambiguity or dynamic occlusion, where temporal context is essential for disambiguation.\\
\indent In this paper, we propose OptiCorNet—an end-to-end trainable sequence-based VPR framework that integrates CNN-based spatial feature extraction, LSTM-based temporal modeling, and a novel Differentiable Sequence Delta (DSD) module. This module performs temporal differencing across sequences to capture dynamic changes, followed by LSTM encoding to learn contextual temporal dynamics. A residual connection allows the network to preserve critical discriminative information across layers. Furthermore, we present a quadruplet loss tailored to VPR scenarios. Unlike the traditional triplet loss, the quadruplet loss incorporates an additional negative sample, which promotes both inter-class dispersion and intra-class compactness within the embedding space. As shown in Figure \ref{fig1}, the proposed OptiCorNet achieves SOTA performance in the sequence-based VPR task.\\
\indent We make the following contributions:
\begin{itemize}
    \item We propose OptiCorNet, an end-to-end framework that jointly performs spatial feature extraction and temporal encoding, enabling direct learning of sequence-level embeddings for accurate and efficient place recognition.
    \item A novel Differentiable Sequence Delta (DSD) module is introduced to capture fine-grained spatiotemporal variations by combining learnable differential weighting with LSTM-based temporal modeling.
    \item To enhance representation robustness, we incorporate a residual connection with a learnable projection, ensuring temporal consistency and improved expressiveness across varying viewpoints and motions.
    \item We design a quadruplet loss function that leverages two negative samples per anchor-positive pair, promoting stronger intra-class compactness and inter-class separation than conventional triplet loss.
    \end{itemize}
\section{Related Work}
\subsection{Visual Place Recognition}
VPR aims to determine whether a given visual observation corresponds to a previously visited location. This field has witnessed a significant evolution from local feature matching toward learning-based paradigms. Existing methods can broadly be categorized into three classes: image-level global descriptor models, sequence-aware approaches, and hybrid architectures that attempt to integrate temporal reasoning into spatial encoding. Global descriptor-based methods primarily operate at the single-image level, extracting compact representations to support scalable retrieval. Early works adopted CNN backbones (e.g., VGG, ResNet) pre-trained on classification datasets and used intermediate activations as descriptors. NetVLAD \cite{arandjelovic2016netvlad} introduced trainable vector aggregation inspired by VLAD encoding, leading to notable gains in robustness and compactness. Subsequent research extended this idea using attention mechanism \cite{vaswani2017attention}, self-supervised contrastive learning \cite{oquab2023dinov2}, and vision transformers \cite{wang2022transvpr}. While effective in controlled settings, these methods lack temporal modeling, which is essential under dynamic conditions.\\
\indent Sequence-based models aim to improve robustness by leveraging sequential consistency. Classical algorithms such as SeqSLAM \cite{milford2012seqslam}, SeqNet \cite{garg2021seqnet} perform appearance-invariant matching via local contrast normalization and sequence alignment. Deep learning variants often adopt RNNs or temporal convolutions to aggregate frame-level embeddings. While these models provide better tolerance to appearance and viewpoint changes, many apply sequence processing as a separate, non-trainable post-hoc step, limiting end-to-end optimization and adaptability.\\
\indent Hybrid and residual-enhanced architectures have recently emerged to unify spatial and temporal learning. Some methods introduce shallow LSTMs atop CNN features, but often treat temporal transitions implicitly. Others utilize differential embeddings \cite{garg2020delta} to capture appearance changes but lack differentiability or integration with learnable dynamics. Moreover, most existing losses are based on triplet \cite{wang2022hybrid} or contrastive objectives \cite{cui2023ccl}, which may not adequately penalize hard negatives or capture fine-grained temporal cues.\\
\indent In contrast, our proposed OptiCorNet introduces a Differentiable Sequence Delta (DSD) module that explicitly captures temporal change vectors via a differentiable delta weighting scheme followed by LSTM-based refinement. The incorporation of residual connections with projection layers ensures dimensional consistency and preserves representational fidelity. Further, our use of a quadruplet loss enhances discriminative learning by simultaneously maximizing inter-class separation and intra-class compactness. To the best of our knowledge, this integration of delta-based reasoning, residual-enhanced sequence modeling, and multi-negative loss is the first to be explored in the sequence-based VPR task in this paper.
\begin{figure*}[http]
    \centering
        \includegraphics[width=0.85\linewidth]{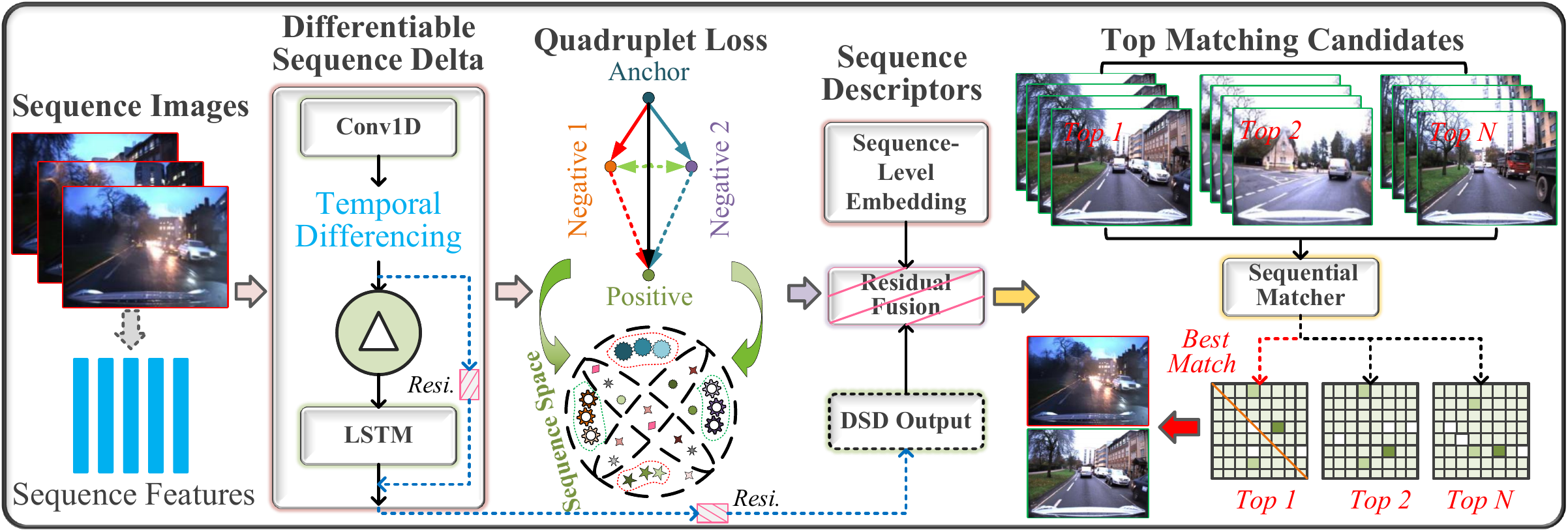}
    \caption{Our proposed OptiCorNet using  DSD module to generate top match hypotheses through a sequential descriptor, followed by quadruplet loss optimized the sequence-level embeddings.}
    \label{fig2}
\end{figure*}
\subsection{Optimization Strategy}
A critical component of learning-based VPR systems lies in the design of effective loss functions that guide the optimization of discriminative and generalizable feature representations. Over the past few years, various loss formulations have been proposed to address the challenges of intra-class variability, inter-class confusion, and environmental changes inherent in real-world place recognition.\\
\indent Triplet loss remains one of the most widely adopted objectives in VPR research. Originally popularized in face recognition \cite{jin2021face} and retrieval tasks \cite{li2025learning}, the triplet loss encourages the distance between an anchor and a positive sample to be smaller than the distance to a negative sample by a fixed margin. Works such as NetVLAD \cite{arandjelovic2016netvlad} and TransVLAD-based VPR \cite{xu2023transvlad} pipelines have incorporated this loss to learn global descriptors under varying conditions. Despite its simplicity, the effectiveness of triplet loss heavily depends on hard negative mining and the choice of margin, which can hinder convergence in large-scale or noisy datasets. Contrastive loss, another classic objective, operates on pairs of samples and seeks to minimize the embedding distance for positive pairs while maximizing it for negatives \cite{du2024probabilistic}, \cite{zhang2024vision}. Although computationally lighter than triplet loss, it lacks a direct mechanism for modeling relative distances among multiple negatives simultaneously, often resulting in less discriminative embeddings when applied to complex sequences.\\
\indent To address these limitations, recent methods have explored multi-negative and hard-negative-aware loss functions. The multi-similarity loss \cite{wang2025focus} and circle loss \cite{liu2024differentiable} adaptively weight positive and negative pairs based on similarity scores, leading to faster convergence and stronger class separation. These formulations have been integrated into recent VPR systems to improve robustness under viewpoint and appearance changes.\\
\indent Our work builds upon the above multi-similarity loss by incorporating a quadruplet loss tailored for short sequence representations, optimizing for both local temporal consistency and global feature discriminability. Combined with our differentiable sequence encoder, this enables robust place matching even in the presence of appearance drift and structural ambiguity.
\subsection{Sequential Descriptors}
Learning effective representations from sequential data has become a central focus in many computer vision tasks, including VPR, action recognition \cite{li2025frame}, video retrieval \cite{liu2025learning}, and trajectory forecasting \cite{guo2024flightbert++}. Rather than treating each frame independently, sequential descriptors aim to encode a temporally coherent segment of observations into a single, compact embedding.\\
\indent The seminal SeqSLAM framework \cite{milford2012seqslam} pioneered the idea of using low-resolution image sequences for place recognition under severe environmental changes. By correlating sequences of image differences, SeqSLAM demonstrated the effectiveness of exploiting temporal coherence even without deep learning. Subsequent deep learning adaptations introduced models such as SeqMatchNet \cite{garg2022seqmatchnet} that extend triplet supervision over short sequences, improving robustness through learned temporal continuity. SeqNet \cite{garg2021seqnet} employs a small 1D CNN on CNN-encoded frame features to learn local temporal structure efficiently, generating compact sequence descriptors. While avoiding RNNs, it only captures short-range dependencies. TimeSformer \cite{mereu2022learning} proposes a detailed taxonomy of techniques using sequential descriptors, highlighting different mechanisms to fuse the information from the individual images. Despite these advances, most methods treat temporal encoding as a post-hoc process, decoupled from spatial feature extraction, limiting the expressiveness of the final embeddings.\\
\indent To address this gap, our work unifies temporal modeling and spatial encoding within a single trainable pipeline. We introduce a DSD module that emphasizes inter-frame changes using a learnable differential weighting scheme, coupled with an LSTM-based encoder. This architecture explicitly encodes temporal dynamics while preserving semantic content via a residual projection, leading to more robust sequence-level embeddings. Furthermore, by integrating these components in an end-to-end framework, we enable direct optimization of the sequence descriptor for the recognition objective.
\section{Proposed Approach}
Here, We present an end-to-end framework for VPR that integrates spatial encoding and temporal dynamics via a Differentiable Sequence Delta (DSD) module with dual residuals. Combined with a quadruplet loss, the model learns robust sequence-level embeddings for reliable place retrieval under challenging conditions. The overall architecture of our proposed OptiCorNet is illustrated in Figure \ref{fig2}. 
\subsection{Differentiable Sequence Delta (DSD)}
\begin{figure}
    \centering
    \includegraphics[width=0.85\linewidth]{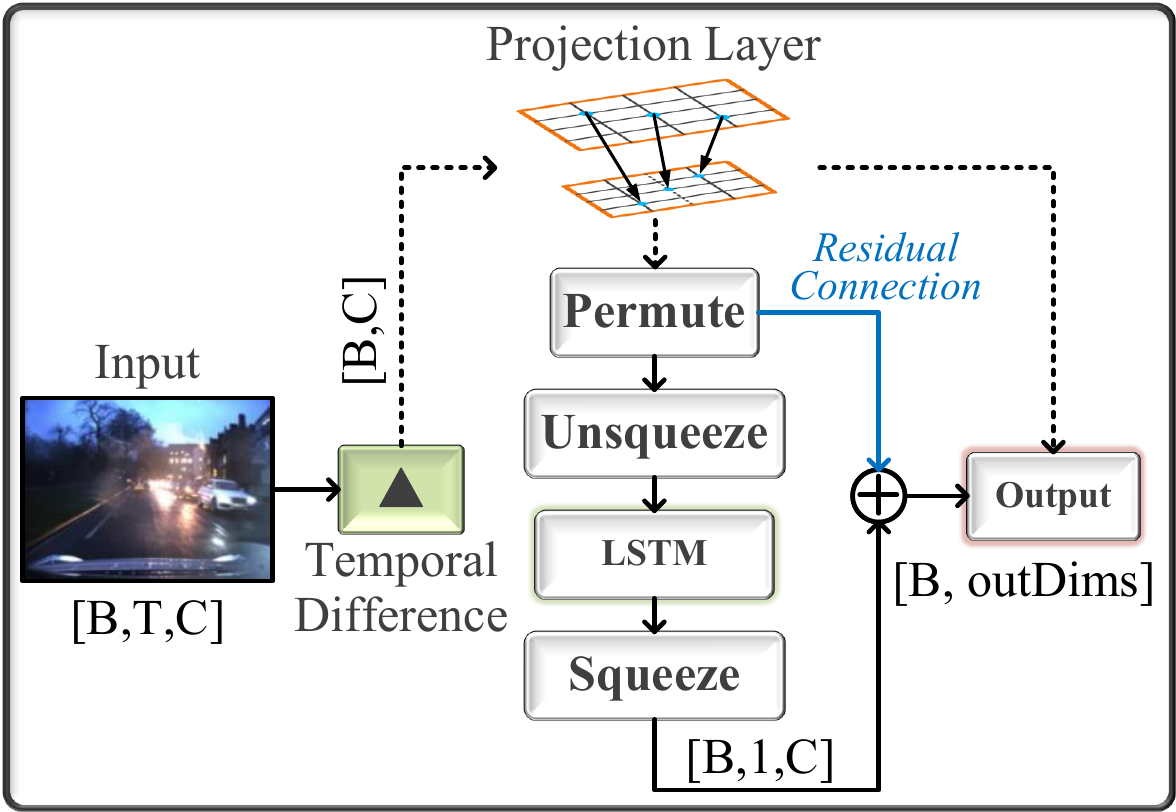}
    \caption{The proposed DSD module. The DSD module captures temporal dynamics by combining learnable frame differencing with LSTM encoding and residual fusion.}
    \label{fig3}
\end{figure}

To effectively capture inter-frame dynamics and enhance temporal reasoning in VPR, we introduce a novel module termed Differentiable Sequence Delta (DSD). This module is designed to extract and encode temporal changes from sequences of feature embeddings through a differentiable, learnable architecture. The overall workflow of DSD is illustrated in Figure.~\ref{fig3}, which includes three main stages: temporal differencing, LSTM-based encoding, and residual projection fusion.\\
\indent Given a sequence of $T$ consecutive frames, their features are first extracted through a deep network backbone, resulting in a 3D tensor $\mathbf{X} \in \mathbb{R}^{B \times T \times C}$, where $B$ is the batch size and $C$ is the feature dimension. To isolate the changes between adjacent frames, we apply a weighted temporal differencing mechanism defined as:
\begin{equation}
\Delta(\mathbf{X}) = \mathbf{X} \cdot \mathbf{w}, \quad \mathbf{w} \in \mathbb{R}^{T}.
\end{equation}
where $\mathbf{w}$ is either a fixed or learnable weight vector with anti-symmetric structure (e.g., $[-1, 0, +1]$). This allows the model to emphasize mid-sequence transitions and suppress redundant edge information. The resulting temporally differenced tensor is denoted as $\mathbf{D} \in \mathbb{R}^{B \times C}$, serving as a compact encoding of inter-frame appearance shifts.\\
\indent To further capture temporal dependencies and contextual cues beyond pairwise differences, we feed $\mathbf{D}$ into a single-layer Long Short-Term Memory (LSTM) unit. LSTM networks are known for their capability to model long-range sequential correlations and for their resistance to vanishing gradients, making them ideal for sequence encoding tasks in dynamic environments. Prior to feeding the LSTM, we reshape $\mathbf{D}$ to match the input shape expected by the recurrent layer:
\begin{equation}
\mathbf{D'} = \text{Unsqueeze}(\text{Permute}(\mathbf{D})) \in \mathbb{R}^{B \times 1 \times C}.
\end{equation}
The LSTM then produces hidden representations $\mathbf{H}$:
\begin{equation}
\mathbf{H} = \text{LSTM}(\mathbf{D'}), \quad \mathbf{H} \in \mathbb{R}^{B \times 1 \times d}.
\end{equation}
where $d$ is the dimensionality of the LSTM output. This captures temporal structure in a compressed representation. The output is then reshaped as:
\begin{equation}
\mathbf{Z} = \text{Squeeze}(\mathbf{H}) \in \mathbb{R}^{B \times d}.
\end{equation}
To preserve the semantic consistency of the original input and stabilize gradient flow, we apply a residual fusion strategy. In cases where the dimensionality of the LSTM output and the temporal difference vector do not match ($d \ne C$), a linear projection layer is used to transform $\mathbf{D}$ into a compatible shape:
\begin{equation}
\mathbf{R} = 
\begin{cases}
\text{Proj}(\mathbf{D}), & \text{if } C \neq d \\
\mathbf{D}. & \text{otherwise}
\end{cases}
\end{equation}
This transformation ensures that the residual path can be effectively added to the LSTM output:
\begin{equation}
\mathbf{F} = \mathbf{Z} + \mathbf{R}.
\end{equation}
Resulting in the final output $\mathbf{F} \in \mathbb{R}^{B \times d}$ of the DSD module. This vector encodes both high-level temporal dependencies and fine-grained appearance shifts, making it particularly suitable for tasks involving dynamic environments and temporal consistency.
\subsection{Discriminative Learning with Quadruplet Loss}
To enforce inter-class separation and intra-class compactness, we build a quadruplet loss function. For each anchor $\mathbf{f}_a$, the loss considers a positive $\mathbf{f}_p$ and two hard negatives $\mathbf{f}_{n1}$, $\mathbf{f}_{n2}$:
\begin{equation}
\mathcal{L}_{quad} = \gamma_1\mathcal{L}_{n1} + \gamma_2\mathcal{L}_{n2}.
\end{equation}
\begin{equation}
\mathcal{L}_{n1} = \max(0, m + \|\mathbf{f}_a - \mathbf{f}_p\|_2^2 - \|\mathbf{f}_a - \mathbf{f}_{n1}\|_2^2).
\end{equation}
\begin{equation}
\mathcal{L}_{n2} = \max(0, m + \|\mathbf{f}_a - \mathbf{f}_p\|_2^2 - \|\mathbf{f}_a - \mathbf{f}_{n2}\|_2^2).
\end{equation}
where $m$ is the margin hyperparameter, $\gamma$ is loss factors. Compared to triplet loss, this formulation encourages more discriminative embedding learning and improves robustness against ambiguous or visually similar negatives.
\begin{figure*}[htbp]
\centering
\subfigure{
\includegraphics[width=1.30in]{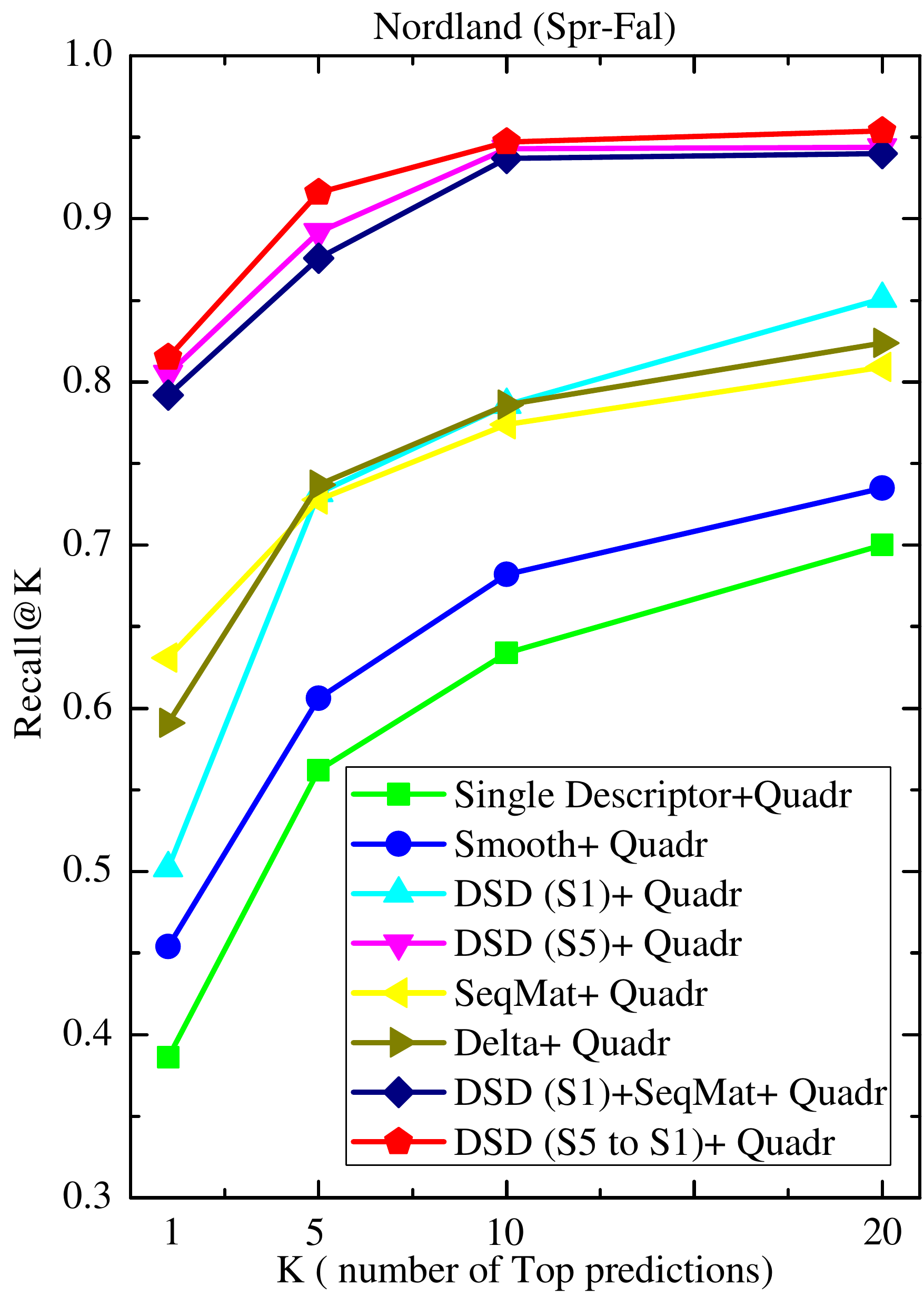}
\label{Fig4-1}
}
\subfigure{
\includegraphics[width=1.30in]{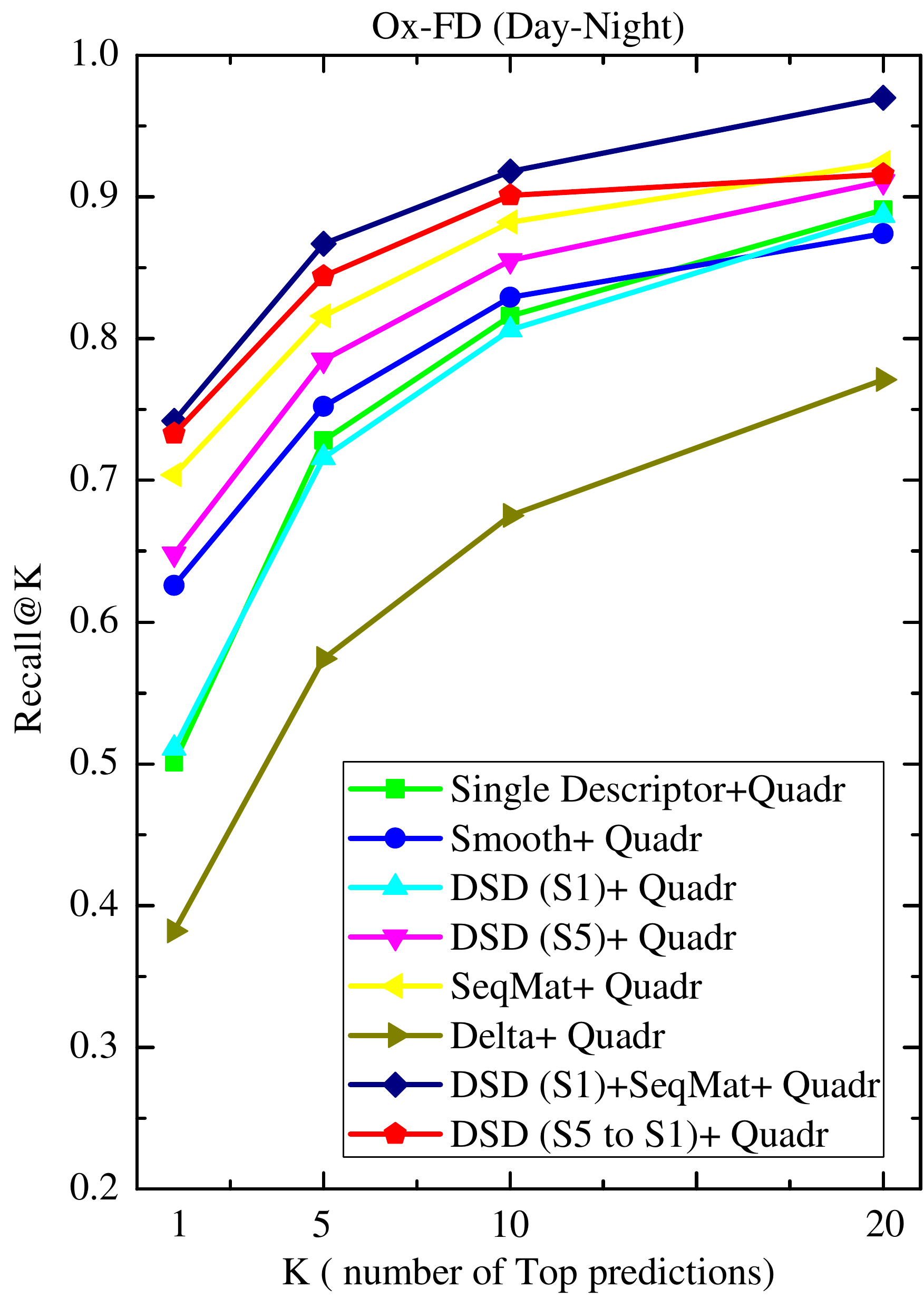}
\label{Fig4-2}
}
\subfigure{
\includegraphics[width=1.30in]{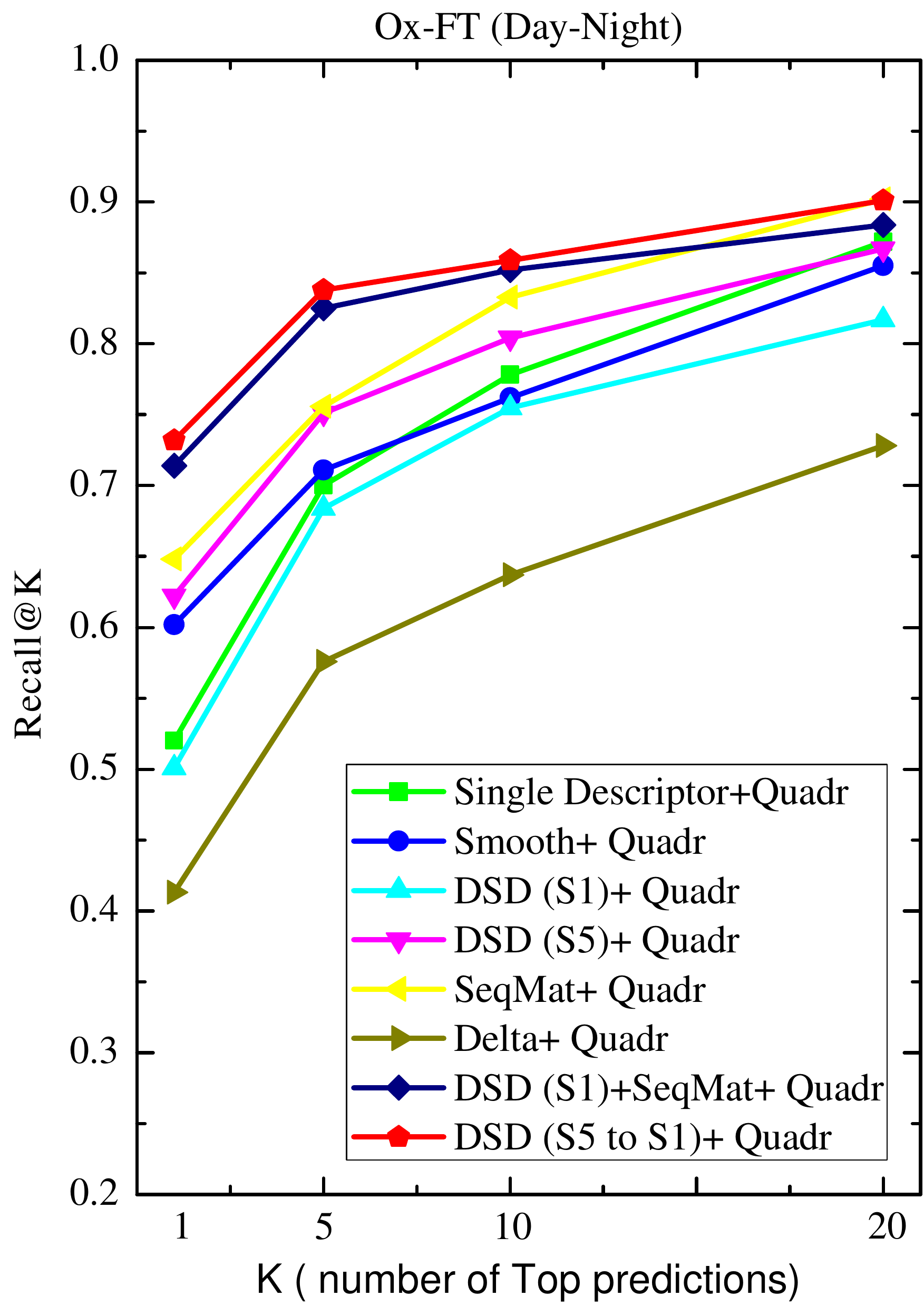}
\label{Fig4-3}
}
\subfigure{
\includegraphics[width=1.30in]{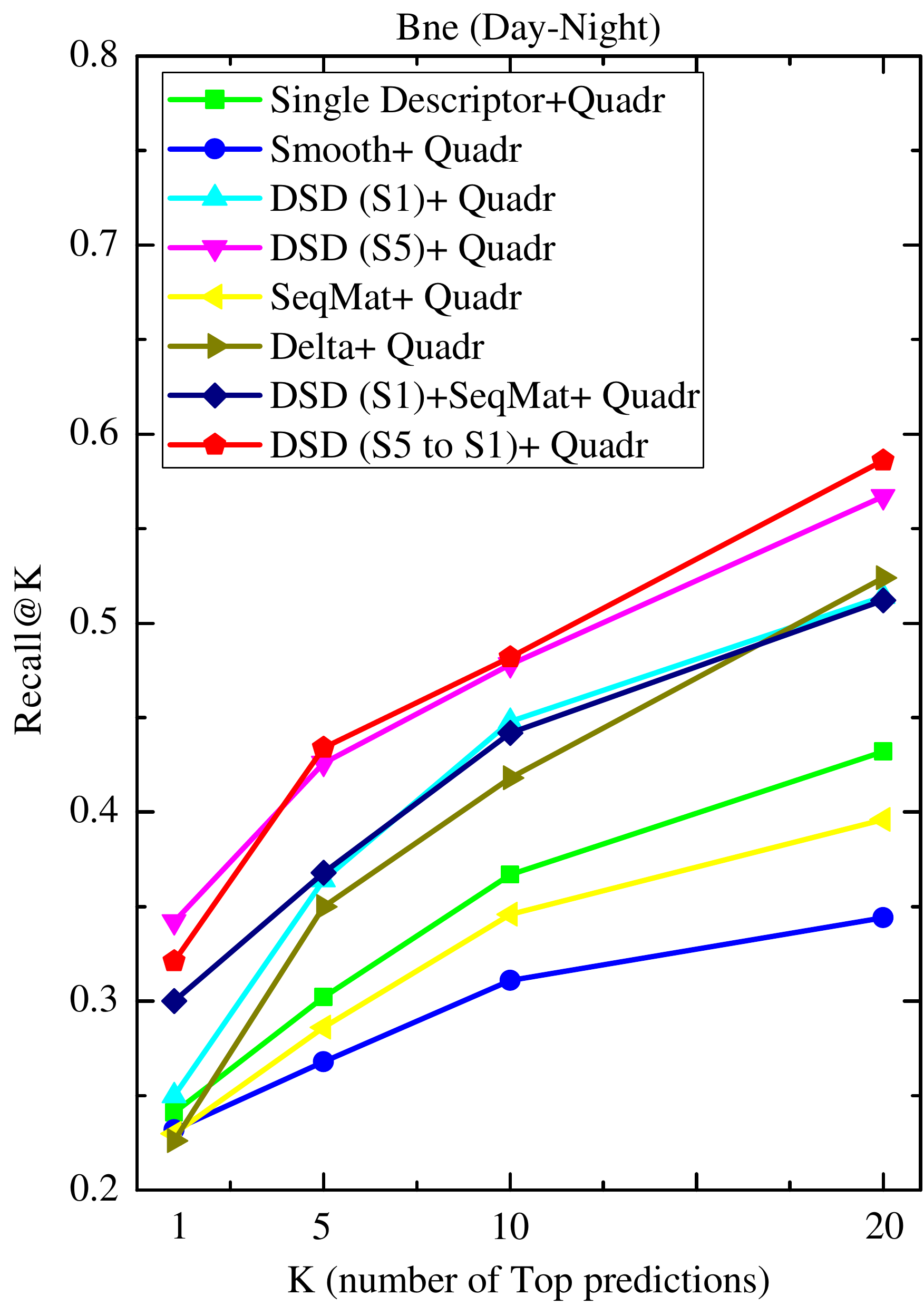}
\label{Fig4-4}
}\\
\subfigure{
\includegraphics[width=1.30in]{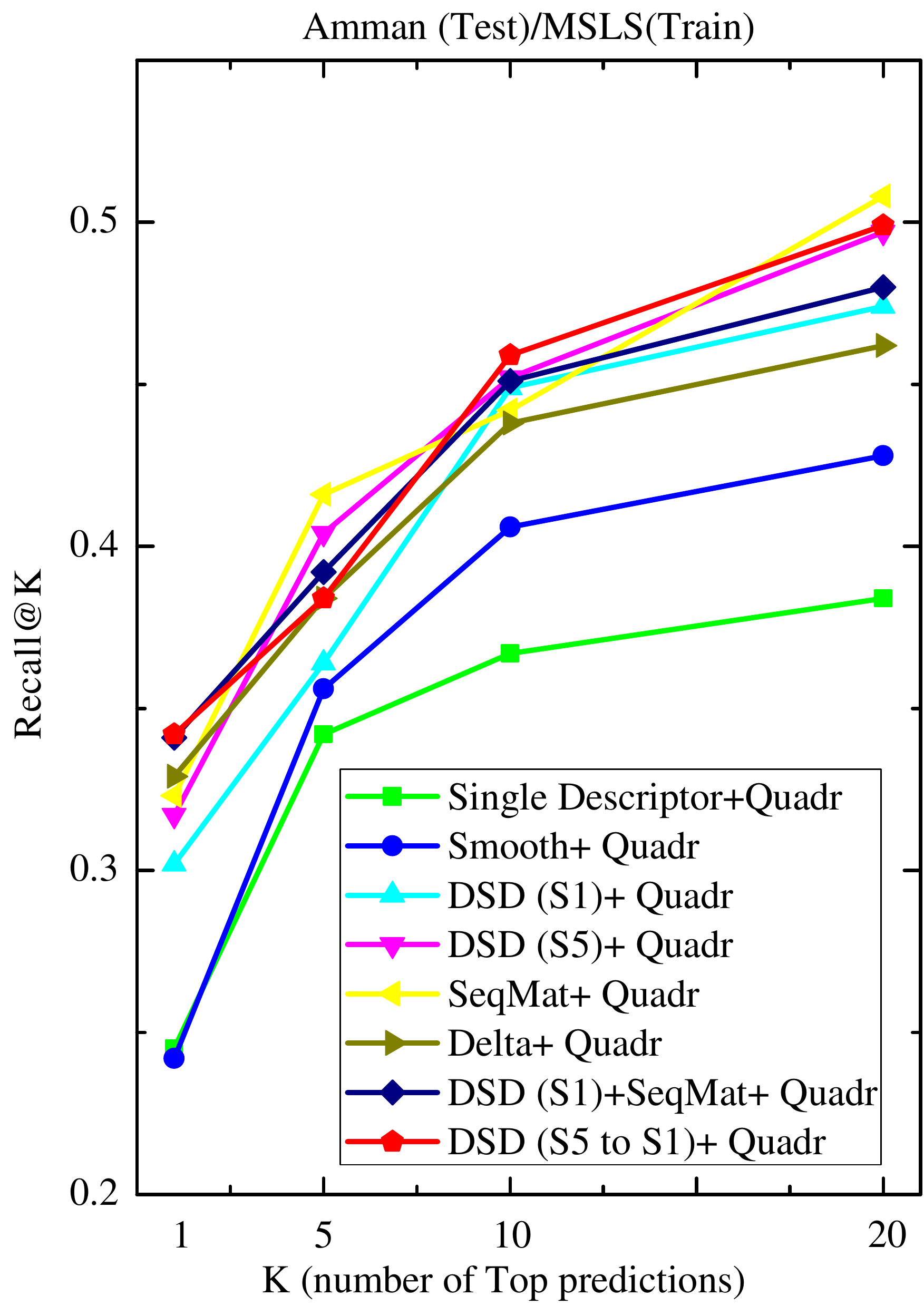}
\label{Fig4-5}
}
\subfigure{
\includegraphics[width=1.30in]{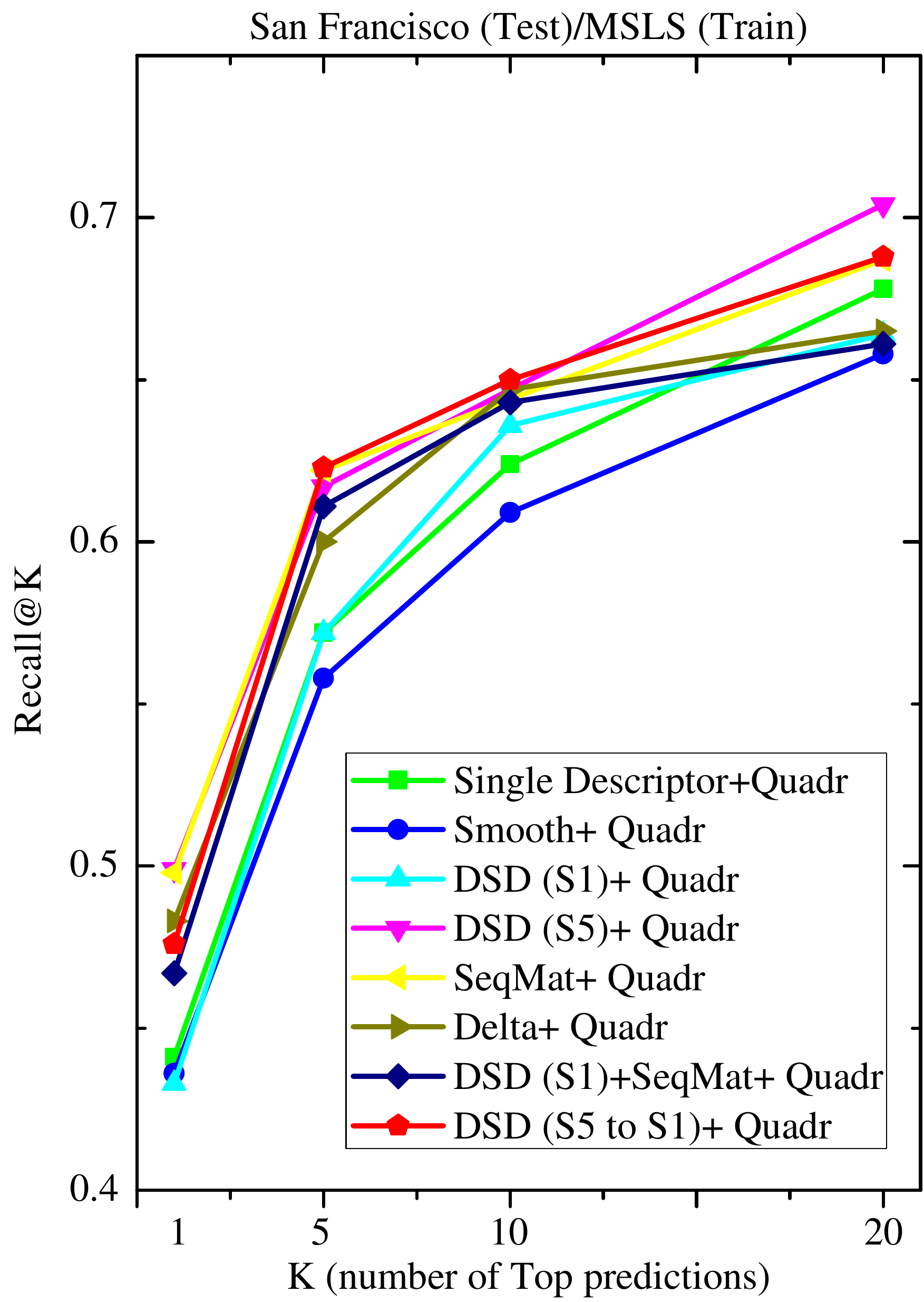}
\label{Fig4-6}
}
\subfigure{
\includegraphics[width=1.30in]{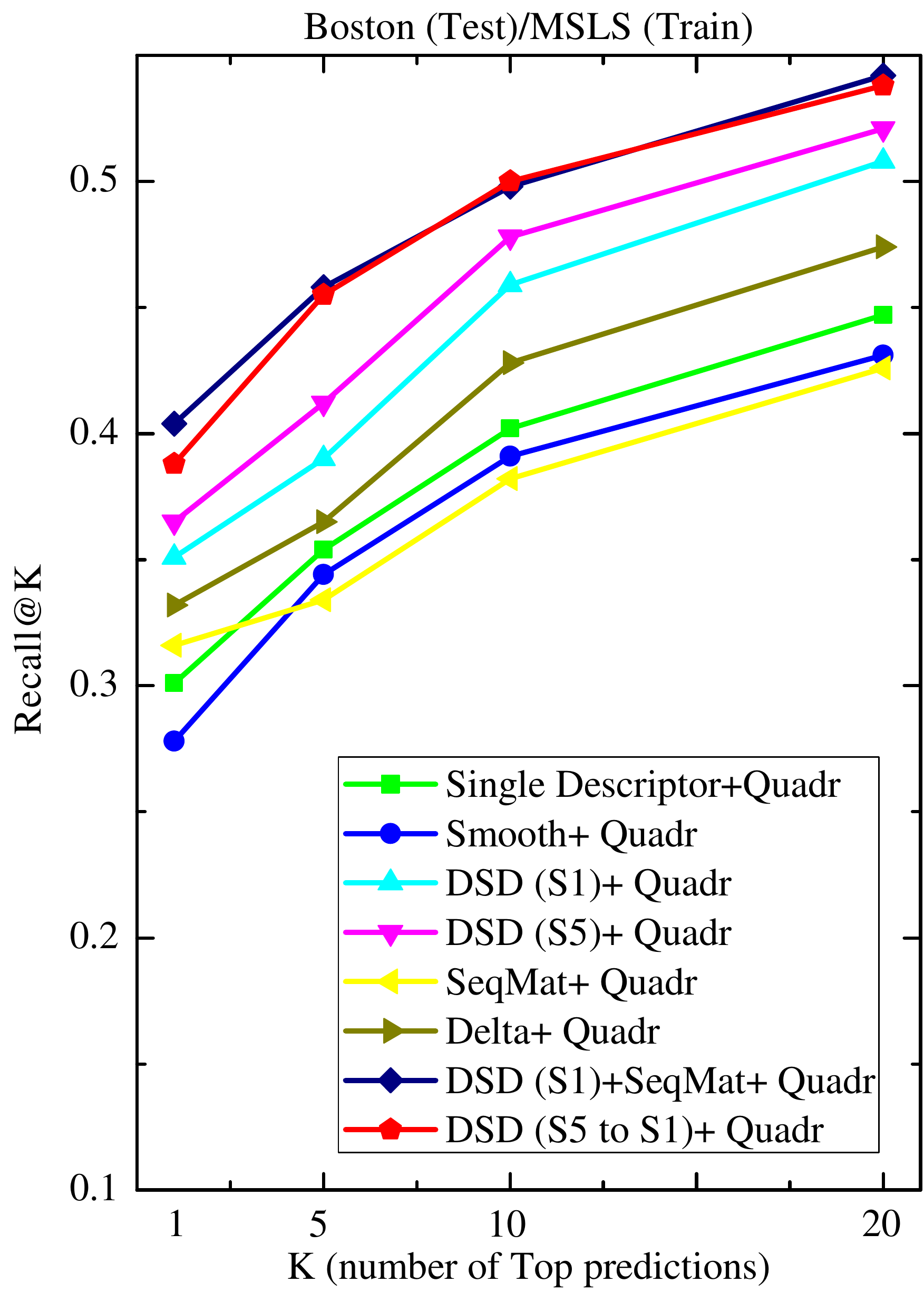}
\label{Fig4-7}
}
\subfigure{
\includegraphics[width=1.30in]{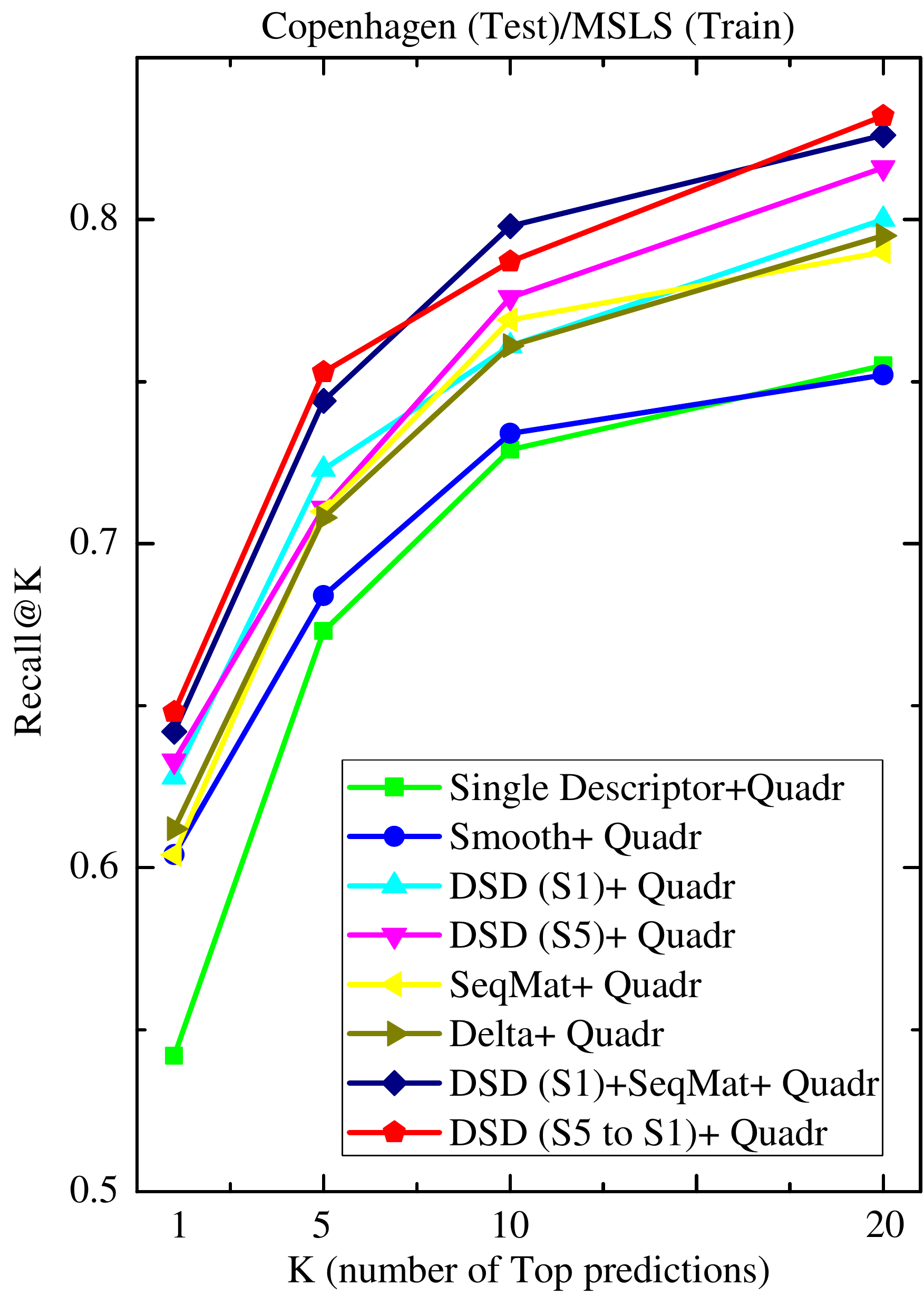}
\label{Fig4-8}
}
\caption{Performance evaluation on some public datasets. We use a localization radius of
 10 meters, 20 meters, and 1 frame, respectively, for Oxford, Brisbane/MSLS, and Nordland datasets. Among them, \textit{A} (Test)/MSLS (Train) means that the model is trained on MSLS and tested on the \textit{A} dataset.}\label{Fig4}
\end{figure*}
\begin{table*}[http]
\centering
\caption{Performance comparison on Nordland dataset. Among them, \textbf{bold} indicates the best result, \underline{underline} indicates the second-best result, and * line indicates the third-best result.}
\begin{tabular}{ccccc}
\hline
\textbf{Method}          & \textbf{Recall@1} & \textbf{Recall@5} & \textbf{Recall@10} & \textbf{Recall@20}   \\ \hline
\textbf{Single Image Descriptors:} \\ 
Conv-AP \cite{ali2022gsv} & 0.38 & 0.54&0.59&0.62 \\
NetVLAD \cite{arandjelovic2016netvlad} & 0.45&0.50&0.72&0.75 \\
Patch-NetVLAD \cite{hausler2021patch} & 0.58&0.74&0.78&0.82 \\
MixVPR \cite{ali2023mixvpr} & 0.76&\underline{0.89}&0.92&\underline{0.95} \\
CosPlace \cite{berton2022rethinking} & 0.54&0.69&0.76&0.83 \\
BoQ \cite{ali2024boq} & 0.70&0.84&0.87&0.89 \\
EigenPlaces \cite{berton2023eigenplaces} & 0.54&0.68&0.74&0.78 \\
SeqNet (S1) \cite{garg2021seqnet} & 0.48&0.69&0.75&0.78 \\ 
SALAD \cite{izquierdo2024optimal}   & 0.76 & \underline{0.89}& - & - \\
EffoVPR (256 DIM) \cite{tzachor2024effovpr} & \underline{0.80}  & -  &  - & - \\
DSD (S1)+ Quadr          & 0.50&0.73&0.78&0.89 \\ \hline
\textbf{Sequential Descriptors:} \\ 
Smoothing \cite{garg2020delta}               & 0.44&0.59&0.67&0.71 \\ 
Delta \cite{garg2020delta}                   & 0.57&0.70&0.76&0.80 \\ 
SeqNet \cite{garg2021seqnet}                  & 0.78&\underline{0.89}&0.92&{0.94*} \\ 
SeqMatchNet \cite{garg2022seqmatchnet}        & 0.66&0.81&0.88&0.91 \\ \hline
\textbf{Sequential Score Aggregation:}\\
Single Descriptor+Quadr  & 0.38&0.56&0.63&0.70 \\ 
Smoothing+ Quadr         & 0.45&0.60&0.68&0.73 \\ 

DSD (S5)+ Quadr          & \underline{0.80}&\underline{0.89}&\textbf{0.94}&\underline{0.95} \\ 
SeqMatch+ Quadr          & 0.63&0.72&0.77&0.80 \\ 
Delta+ Quadr             & 0.59&0.73&0.78&0.82 \\ 
DSD (S1)+SeqMat+   Quadr & {0.79*}&{0.87*}&\underline{0.93}&{0.94*} \\ 
DSD (S5 to S1)+   Quadr  & \textbf{0.81}&\textbf{0.91}&\textbf{0.94}&\textbf{0.96} \\ \hline
\end{tabular}
\label{table1}
\end{table*}
\subsection{Sequence-Level Retrieval}
To robustly determine the final match between query and reference trajectories in challenging environments, we employ a hierarchical sequence-level retrieval strategy inspired by HVPR \cite{garg2021seqnet} that exploits both globally encoded sequence descriptors and locally refined frame-level alignment. Our framework first generates a summary-level embedding for each sequence by encoding a set of frame-wise features through the proposed DSD module. These high-level descriptors $\mathbf{S}_{L_d}$ reside in a compact $D$-dimensional space and are used to shortlist candidate matches based on Euclidean distance.\\
\indent Given a query sequence $q$, we compute its global sequence embedding $\mathbf{S}_{L_d}^q$ and compare it against all reference embeddings $\mathbf{S}_{L_d}^j$ in the database using the following similarity metric:
\begin{equation}
p_{ij} = \|\mathbf{S}_{L_d}^q - \mathbf{S}_{L_d}^j\|_2, \quad \forall j \in \mathcal{D},
\end{equation}
where $\mathcal{D}$ denotes the reference database. This initial ranking identifies the top $K$ candidate locations $R_i$ with the smallest distances. These candidates are further refined through a local sequence matching procedure based on learned single-image descriptors $\mathbf{S}_1$.\\
\indent To assess local alignment, we perform a fixed-length sequence comparison using a simplified SeqSLAM-style scoring function without velocity variation:
\begin{equation}
q_{ik} = \sum_{t=0}^{L_m - 1} \|\mathbf{S}_1^{i-t} - \mathbf{S}_1^{k-t}\|_2, \quad \forall k \in R_i,
\end{equation}
where $L_m$ denotes the local sequence length, and $\mathbf{S}_1^{i-t}$ and $\mathbf{S}_1^{k-t}$ are the learned single-image descriptors from query and candidate sequences, respectively. This sequential comparison assumes temporal consistency and one-to-one correspondence, minimizing alignment distortion caused by frame-wise misregistrations.\\
\indent Finally, we select the candidate $k^*$ that minimizes the alignment score:
\begin{equation}
k^* = \arg\min_{k \in R_i} q_{ik}.
\end{equation}
\indent This hierarchical process enhances robustness by leveraging coarse-to-fine temporal reasoning: the global embedding narrows down the search space, while the local descriptor sequence ensures precise alignment. Such a design allows our model to handle significant appearance changes.
\section{Experimental Setup}
\subsection{Datasets and Evaluation Metrics}
\subsubsection{Datasets}
To rigorously assess the robustness and generalization of our proposed sequence-based visual place recognition framework, we evaluate it across publicly available datasets, each reflecting distinct real-world challenges. These datasets encompass large-scale variations in lighting, seasonal conditions, urban complexity, and geographic diversity. Specifically, we utilize: (1) the Oxford and Brisbane urban datasets, which capture day-to-night traversals under varying illumination and dynamic traffic conditions, with each city providing stereo sequences over 10 km and 25K–30K frames; (2) the Nordland railway dataset, which includes a 728 km train journey recorded across four seasons with consistent viewpoints, allowing us to evaluate long-term seasonal robustness using standard Summer–Winter splits; (3) the MSLS dataset, offering geo-tagged street-view imagery from 30 global cities with heterogeneous weather, time-of-day, and camera setups—ideal for benchmarking cross-city generalization. 
\subsubsection{Evaluation Metrics}
VPR methods are often integrated into localization pipelines to generate coarse location priors, Since subsequent modules typically achieve high precision in refining poses once a correct candidate, the primary requirement for VPR systems is to ensure high Recall rather than top-1 accuracy, which motivates the use of Recall@K as a core performance metric.\\
\indent Recall@K measures the proportion of query images for which a correct database match exists within the top-$K$ retrievals. A match is considered correct if its place lies within a pre-defined localization radius $R$ from the ground-truth location. Formally, Recall@K is computed as:
\begin{equation}
\text{Recall@}K = \frac{1}{N} \sum_{i=1}^{N} \mathbb{I} \left( \min_{j \in \text{TopK}_i} d(q_i, r_j) \leq R \right)
\label{eq:recallatk}
\end{equation}
where $N$ denotes the total number of queries, $q_i$ is the ground-truth position of the $i^{th}$ query, $\text{TopK}_i$ is the set of indices corresponding to the top-$K$ retrieved database candidates, and $r_j$ is the position of the $j^{th}$ candidate. The function $d(\cdot, \cdot)$ denotes a suitable distance metric (e.g., Euclidean distance or frame difference), and $\mathbb{I}(\cdot)$ is the indicator function.\\
\indent To account for differences across datasets, we follow prior works and adopt dataset-specific thresholds for the localization radius $R$: 10 meters for the Oxford dataset, 20 meters for the Brisbane and MSLS datasets, and 1 frame for the Nordland dataset. This metric provides a reliable measure of a VPR model’s ability to propose viable candidates for downstream pose refinement, which is critical for robust localization performance.
\subsection{Implementation Details}
We train our model using a quadruplet loss framework, where each training sample consists of a query image, a positive, a hard negative, and an additional negative to improve feature separation. The mini-batch size is set to $N=32$ quadruplets, with each tuple composed of $L=12$ images. The sequence length is set to $\ell=5$, and the temporal filtering kernel has width $w=3$. We use NetVLAD \cite{arandjelovic2016netvlad} as the base descriptor extractor with output dimensionality $D$, and apply sequence-aware pooling to generate compact embeddings. Optimization is carried out using SGD with a momentum of $0.9$, weight decay of $1\mathrm{e}^{-3}$, and an initial learning rate of $1\mathrm{e}^{-4}$, which is decayed by a factor of $0.5$ every $50$ epochs. The model is trained for $200$ epochs. Cache refreshing for hard negative mining is optionally triggered every $R$ queries. Model checkpoints are saved at each evaluation step, and the one with the highest Recall on the validation set is used for final testing.
\subsection{Evaluation for Differentiable Sequence Delta (DSD)}
Figure \ref{Fig4} presents Recall@K performance across diverse VPR scenarios. The proposed DSD(S1)+Quadruple loss (Quadr) method consistently achieves the highest accuracy across datasets, including seasonal (Nordland), illumination (Oxford/Brisbane), and domain-shifted (MSLS cross-city) conditions. Compared to single descriptors, smoothing, or delta baselines, DSD demonstrates superior robustness and generalization, particularly in challenging settings with strong perceptual aliasing. The integration of LSTM-based modeling and SeqMatch ranking enhances discriminative capability while reducing false matches. Overall, the DSD pipeline offers reliable sequence-level retrieval with strong cross-domain performance.
\subsection{Comparison with other Methods}
As shown in Table \ref{table1}, our proposed DSD module demonstrates the superiority under challenging seasonal variations in the Nordland dataset. Although image-level descriptors show strong robustness in extreme environments, our proposed sequence-based matching can also achieve a similar effect. Notably, our approach, DSD (S5 to S1) + Quadr, achieves the highest performance across all metrics, validating the effectiveness of combining learnable differencing, residual fusion, and quadruplet loss. Compared to other sequence-based baselines like SeqNet and SeqMatchNet, DSD-based variants offer more discriminative and robust representations, especially under extreme appearance shifts.
\subsection{Ablation Study}
The ablation study, as shown in Table \ref{table2}, underscores the effectiveness of each component within OptiCorNet. Removing the LSTM module slightly reduces Recall@1 (from 0.81 to 0.80), confirming the benefit of temporal modeling for capturing sequential dependencies. Excluding residual connections or fusion results in further degradation, especially at higher recall thresholds, highlighting their role in preserving semantic integrity and stabilizing feature learning. Notably, omitting the differencing operation leads to a substantial drop, emphasizing the importance of capturing inter-frame dynamics via the DSD module. Finally, the use of quadruplet loss consistently outperforms triplet-based supervision, demonstrating improved discriminative capacity of the learned embeddings. 
\begin{table}[http]
\centering
\caption{Ablation study of key components of the proposed OptiCorNet on Nordland dataset.}
\begin{tabular}{cc}
\hline
\textbf{Method}           & \textbf{Recall@1/5/10/20}    \\ \hline
w/o LSTM                  & 0.80/0.89/0.91/0.93 \\ 
w/o residual connection   & 0.78/0.87/0.90/0.92 \\ 
w/o residual fusion       & 0.78/0.85/0.88/0.92 \\ 
w/o differencing          & 0.72/0.78/0.82/0.83 \\ \hline
OptiCorNet (Triplet loss) & 0.79/0.90/0.93/0.95 \\ 
OptiCorNet                & \textbf{0.81}/\textbf{0.91}/\textbf{0.94}/\textbf{0.96} \\ \hline
\end{tabular}
\label{table2}
\end{table}
\subsection{Evaluation for Different Retrieval Strategies}
Table \ref{table3} demonstrates a clear trend: increasing sequence length from S1 (single image) to S5 (stride-5 sequence) improves retrieval accuracy, with Recall@1 rising from 0.50 to 0.80. The hierarchical strategy (S5 to S1) achieves the best performance (Recall@1 = 0.81), benefiting from coarse-to-fine filtering while maintaining reasonable computational cost ($67^{\pm5}$ms). While S1 and S2 are faster, their limited temporal context results in significantly lower accuracy. 
\begin{table}[http]
\centering
\caption{Time Cost based on Different Retrieval Strategies}
\begin{tabular}{ccc}
\hline
\textbf{Method}       & \textbf{Recall@1/5/10/20}    & \textbf{Time (ms)} $\downarrow$ \\ \hline
S1                    & 0.50/0.73/0.78/0.89 & \textbf{$38^{\pm5}$}                          \\ 
S2                    & 0.55/0.75/0.80/0.90 & $42^{\pm5}$                          \\ 
S3                    & 0.66/0.78/0.84/0.91 & $49^{\pm5}$                           \\ 
S4                    & 0.75/0.81/0.88/0.93 & $54^{\pm5}$                           \\ \hline
S5       & 0.80/0.89/0.94/0.95 & $56^{\pm5}$                          \\ 
S5 to S1 & \textbf{0.81}/\textbf{0.91}/\textbf{0.94}/\textbf{0.96} & $67^{\pm5}$                          \\ \hline
\end{tabular}
\label{table3}
\end{table}
\subsection{Robust Test}
\begin{figure}
    \centering
    \includegraphics[width=0.80\linewidth]{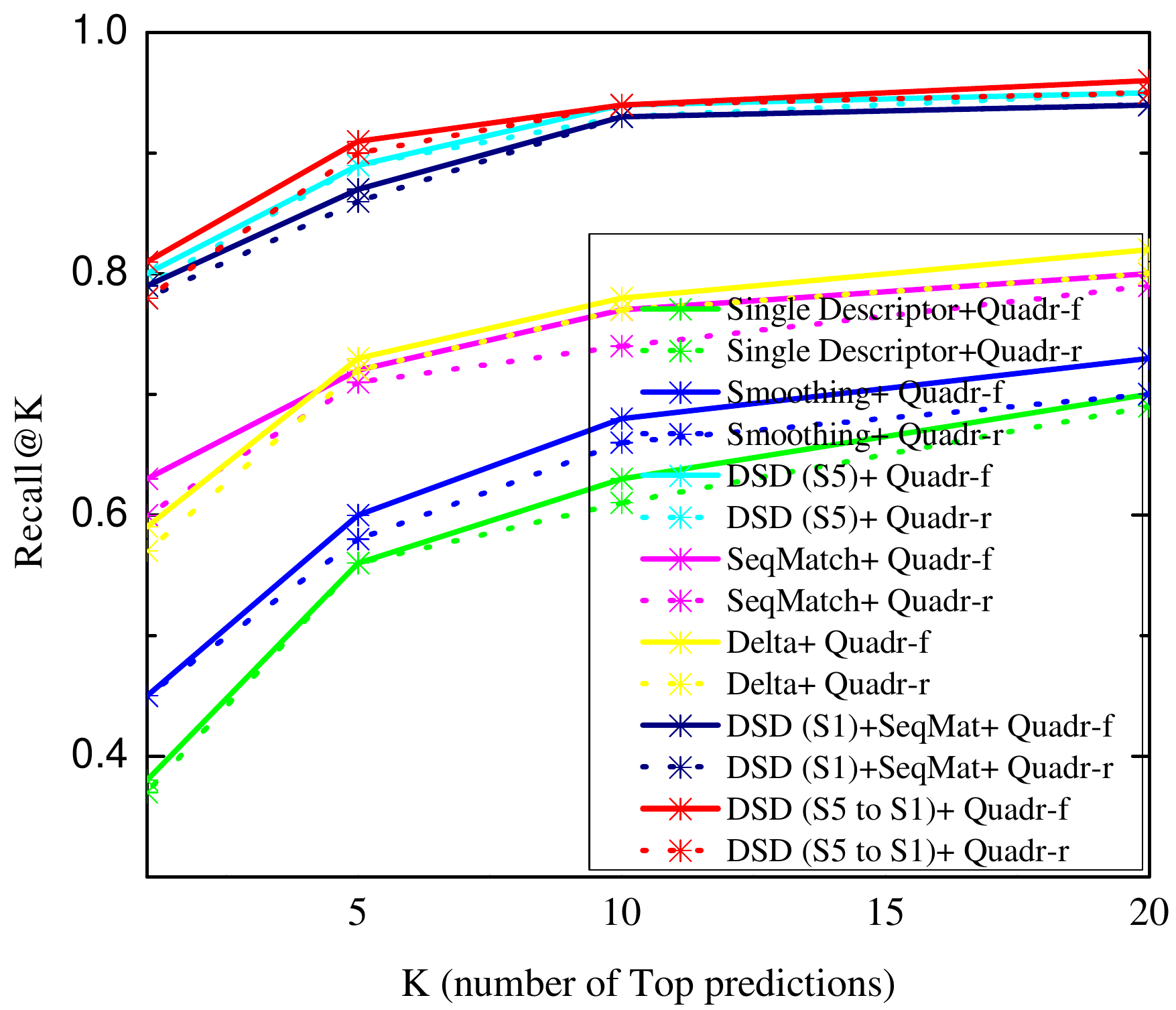}
    \caption{Robust evaluation on Nordland dataset. The solid line represents the forward search, while the dotted line represents the corresponding reverse search.}
    \label{fig5}
\end{figure}

To verify the proposed method's robustness under different test methods, we perform a reverse retrieval experiment on the Nordland dataset. In this experiment, the processing order of the reference database is opposite to that of the query process, while the query process itself remains unchanged. The results presented in Figure \ref{fig5} indicate that the DSD module does not significantly alter the outcomes of the robustness test. However, the solutions that do not incorporate the DSD module—namely, Single Descriptor + Quadr, and slightly lower performance, which suggests that the DSD module we proposed positively contributes to maintaining robustness.
\section{Conclusions}
This paper proposes a unified and differentiable sequence-based visual place recognition framework that combines a novel Differentiable Sequence Delta (DSD) module, temporal modeling via LSTM, and residual fusion for robust spatiotemporal representation learning. Through extensive experiments across challenging benchmarks, our approach consistently outperformed state-of-the-art baselines in both accuracy and generalization. Notably, our method achieved a Recall@1 of 0.81 on the Nordland dataset, surpassing previous sequential methods such as SeqNet and Delta. Ablation studies further validated the importance of key component: removing LSTM or DSD notably degraded performance, underscoring their complementary roles in encoding temporal coherence. Moreover, the proposed DSD (S5 to S1) hierarchical strategy demonstrated superior accuracy while maintaining efficient inference time.\\
\indent Although sequence-based VPR methods offer superior efficiency and scalability, their performance still lags behind deep re-ranking approaches under extreme appearance changes. However, sequence-based retrieval remains an attractive choice for real-world autonomous systems due to its lower computational overhead and higher inference efficiency. Future research should focus on enhancing robustness through hybrid models that integrate semantic reasoning and temporal modeling, while preserving real-time inference capabilities crucial for autonomous navigation.

\bibliography{aaai2026}

\end{document}